\documentclass[10pt, conference]{IEEEtran}
\usepackage{times}
\usepackage[utf8]{inputenc}
\usepackage{graphicx}
\usepackage{amsmath}
\usepackage{amssymb}
\usepackage{booktabs}
\usepackage{multirow}
\usepackage{subcaption}
\usepackage{geometry}
\title{CAP-GAN: Towards Adversarial Robustness with Cycle-consistent Attentional Purification}
\author{
\IEEEauthorblockN{Mingu Kang}
\IEEEauthorblockA{\textit{School of Computing} \\
\textit{KAIST}\\
Daejeon, South Korea \\
mingu.kang@kaist.ac.kr}
\and 
\IEEEauthorblockN{Trung Quang Tran}
\IEEEauthorblockA{\textit{School of Computing} \\
\textit{KAIST}\\
Daejeon, South Korea \\
trungtq2019@kaist.ac.kr}
\and
\IEEEauthorblockN{Seungju Cho}
\IEEEauthorblockA{\textit{School of Computing} \\
\textit{KAIST}\\
Daejeon, South Korea \\
joyga@kaist.ac.kr}
\and
\IEEEauthorblockN{Daeyoung Kim}
\IEEEauthorblockA{\textit{School of Computing} \\
\textit{KAIST}\\
Daejeon, South Korea \\
kimd@kaist.ac.kr}
}
\begin{document}

\maketitle
\begin{abstract}
Adversarial attack is aimed at fooling a target classifier with imperceptible perturbation. Adversarial examples, which are carefully crafted with a malicious purpose, can lead to erroneous predictions, resulting in catastrophic accidents. To mitigate the effect of adversarial attacks, we propose a novel purification model called CAP-GAN. CAP-GAN considers the idea of pixel-level and feature-level consistency to achieve reasonable purification under cycle-consistent learning. Specifically, we utilize a guided attention module and knowledge distillation to convey meaningful information to the purification model. Once the model is fully trained, inputs are projected into the purification model and transformed into clean-like images. We vary the capacity of the adversary to argue the robustness against various types of attack strategies. On CIFAR-10 dataset, CAP-GAN outperforms other pre-processing based defenses under both black-box and white-box settings.
\end{abstract}
\section{Introduction}
\label{Introduction}
With Deep Neural Networks (DNNs) advancement, utilizing DNNs in various applications is in full swing.
Among them, the image classification task aims to train DNNs to classify given images correctly.
DNNs have shown outstanding successes in the image classification task and are being studied focusing on improving the accuracy \cite{he2016deep,szegedy2016inception}. 
However, DNNs have been proved to be vulnerable against imperceptible noise, which is carefully crafted by an adversary, called \textit{adversarial attacks} \cite{goodfellow2014explaining}. 
Such a weak point of DNNs raises security concerns in that the machine cannot entirely substitute for the human.
One of the main concerns is that adversarial examples force DNNs to behave unexpectedly, despite the fact that adversarial examples are seemingly similar to the original one. 

To combat adversarial attacks, two kinds of defense mechanisms have been mainly proposed: \textit{adversarial training} \cite{goodfellow2014explaining,szegedy2013intriguing,madry2017towards}, where a target model is directly re-trained with adversarial examples to obtain the adversarial robustness, and \textit{pre-processing}, where another model is designed to transform inputs from malicious to clean-like inputs.
Although adversarial training shows outstanding robustness across various types of attacks, it generally requires extreme training resources and significantly deteriorates generalization ability. 
Unlike adversarial training, pre-processing comes at a relatively low cost and shows better generalization.
Thus, pre-processing is worth further being explored in that it is more applicable than adversarial training.
When it comes to pre-processing, various defenses have been proposed to mitigate the effect of adversarial attacks \cite{song2017pixeldefend,liao2018defense,xie2019feature,guo2017countering,guo2017countering,samangouei2018defense,xiao2020one}. 
They came up with the denoising function to remove the adversarial perturbation from inputs.
\begin{figure}[h!]
\begin{center}
\includegraphics[width =8cm]{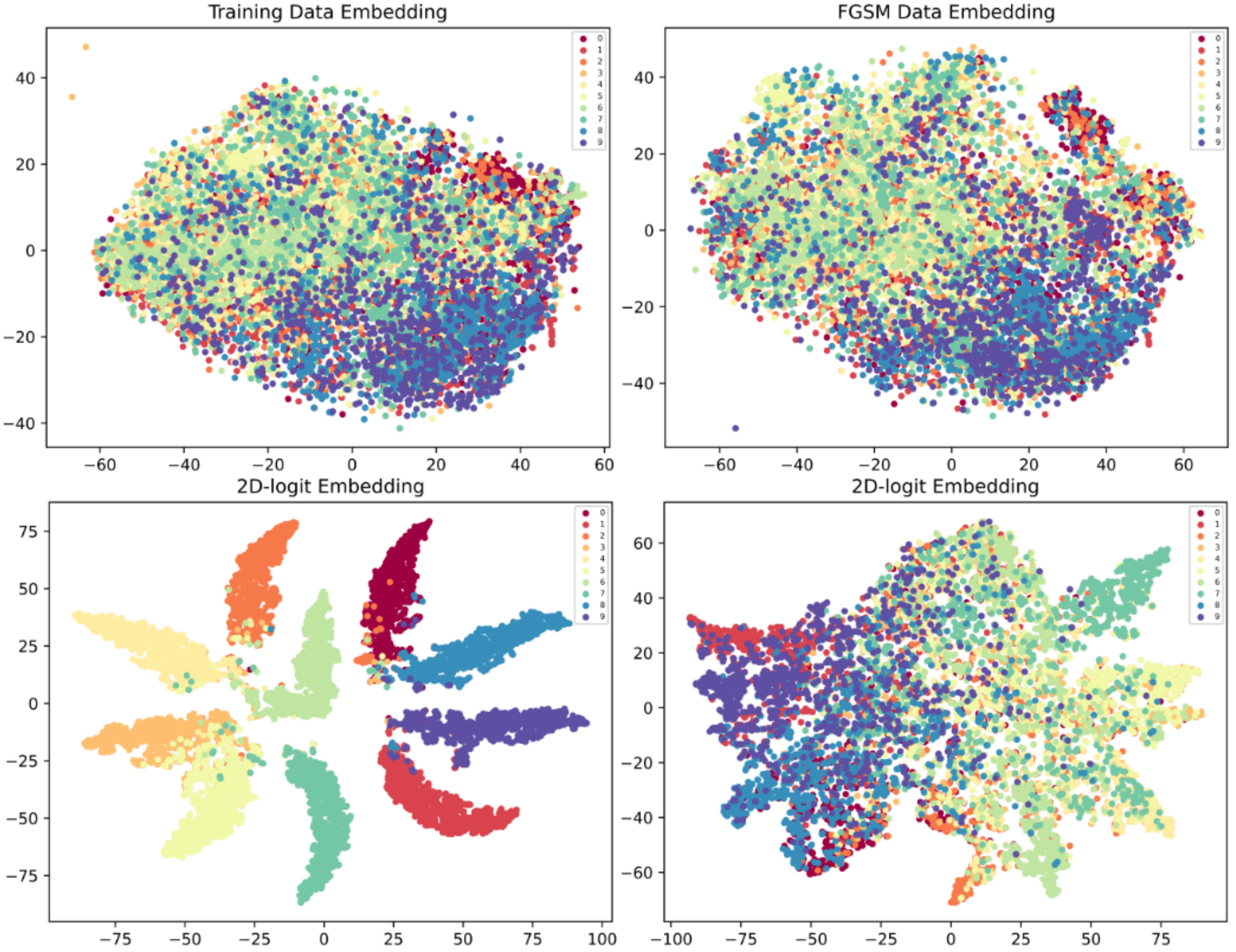}
\end{center}
\caption{The first row represents t-SNE visualization of CIFAR-10 with $x^{i} \in \mathbb{R}^{H \times W \times C} $ and corresponding FGSM \cite{goodfellow2014explaining} adversarial data with a large magnitude of perturbation $\epsilon$ =16, $x^{i}_{adv} \in \mathbb{R}^{H\times W\times C}$(left and right on the first row respectively), which can be viewed as differences in pixel-level distribution.
The second row represents differences in feature-level distribution extracted by $f$ : $f(x^{i})$ and $f(x^{i}_{adv})$. 
}
\label{tsne_viz}
\end{figure}

Intuitively, a way to mitigate the effect of adversarial attacks is to get rid of the adversarial perturbation in a pixel-level approach.
Finding a perturbation for each sample and disentangling it is plausible at first glance. 
Unfortunately, it is impossible to find the optimal distribution of the perturbation, which can cover all possible attacks, due to the high complexity of DNNs. 
Considering one of the main spirits of adversarial examples, $x_{adv} = x + \delta$, where perturbation $\delta$ tends to be imperceptible to satisfy the alignment of human perception with a small budget of the perturbation, pixel-level purification might not be suitable to mitigate the effect of adversarial attacks.
Even if we can achieve proper pixel-level alignment, it would not bring feature-level alignment  \cite{liao2018defense,hoffman2018cycada}.

Specifically, given an original data $x$ and corresponding adversarial data $x_{adv}$,  their logit can be presented as $f(x)$ and $f(x_{adv})$ respectively, where $f$ is a pre-trained network.
We call the logit a feature-level vector of the input extracted from $f$.
We visualize pixel-level and feature-level distribution to compare how adversarial attack affects both spaces.
In Figure \ref{tsne_viz}, when it comes to the visual domain shift $x,x_{adv} \in \mathbb{R}^{H \times W \times C}$, it is hard to observe discernible changes that stand out, whereas the feature-level space $f(x), f(x_{adv}) \in \mathbb{R}^{K}$ has been significantly hurt by the adversarial perturbation, where $K$ is the number of classes.
This result indicates that adversarial examples would focus more on becoming counterparts in the high-level representation space.
The more sophisticated attacks are, the deeper this tendency will be.


In spite of the characteristics of adversarial attack, previous defenses only considered the pixel-level distribution \cite{song2017pixeldefend,shen2017ape,samangouei2018defense}, just with a hope that aligning the pixel-level distribution can bring the feature-level alignment; given defense mechanism $\mathcal{T}$, it will satisfy a term, where $f(\mathcal{T}(x_{adv})) \approx f(x) $ if $\mathcal{T}(x_{adv}) \approx x$.
Instead of the implicit hope, recent defenses start turning their attention to adopt feature-level objective functions to ensure that generated images are tailored for the end-task purpose, e.g., classification \cite{xiao2020one,xie2019feature,liao2018defense}.
Equipped with these perspectives, we envision the purification model $\mathcal{T}$, which eliminates adversarial perturbations by leveraging pixel-level and  feature-level approach.

We propose a novel way to mitigate the effect of the adversarial perturbation via input transformation.
Given the pre-trained model $f$, we train the purification model to improve the robustness of $f$ for the image classification.
We utilize GAN \cite{goodfellow2014generative} training procedure to purify adversarial inputs. 
Concretely, CAP-GAN adopts CycleGAN \cite{zhu2017unpaired} training scheme, where a forward pass of the model has cycle-consistency to satisfy the pixel-level consistency. 
For the feature-level consistency, we introduce explicit terms into $\mathcal{T}$ by distilling the knowledge of the pre-trained model $f$.
To guide the model to focus on meaningful regions, we use an attention module inspired by previous researches \cite{kim2019u,zhou2016learning}.
We found that introducing feature-level terms enhances the robustness fairly compared to the model that only considers the pixel-level alignment.
Moreover, the model can be further improved under cycle-consistent learning.

Instead of a simple dataset like MNIST, we evaluate our CAP-GAN on CIFAR-10 across various types of adversarial attacks.
In addition, we carry out an adaptive attack in which the adversary exploits an internal weakness of the proposed method.
Our model outperforms other pre-processing based defenses under both black-box and white-box settings, according to the experimental results.
To sum it up, our main contributions are listed as follows:
\begin{itemize}
\item To resist adversarial attacks, we propose a new pre-processing method, called CAP-GAN, considering both pixel-level and feature-level alignment under cycle-consistent learning.
\item Especially, we apply guided attention module and knowledge distillation to achieve feature-level consistency.
\item Our model outperforms other pre-processing methods and achieves promising robustness against various attack strategies under both black-box and white-box settings.
\end{itemize}

\section{Related Work}
\subsection{Adversarial Attacks}
Adversarial examples can be crafted by adjusting loss values of given images. After finding adversarial examples by \cite{szegedy2013intriguing}, many researchers have developed various algorithms for crafting adversarial examples \cite{goodfellow2014explaining, madry2017towards, carlini2017towards, uesato2018adversarial}. Fast Sign Gradient Method (FGSM) \cite{goodfellow2014explaining}, and Projected Gradient Descent (PGD) algorithm \cite{madry2017towards} are typically used for evaluating the robustness of the model. Both algorithms use the gradient of the loss function with respect to the input images using the following formula.

\begin{equation}
\nonumber
X_0 = X, X_{i+1} = \Pi(X_i + \alpha  \cdot  sign(\triangledown_{X} l(X_{t},y_{true})))
\label{adv_attack}
\end{equation}
Here $\alpha$ is a hyper-parameter, and $\Pi$ is a projection function for adjusting perturbation size. 
In this formula, it is FGSM algorithm if $i$ is 1, and PGD uses more than 1. 
Carlini and Wagner (CW) attack approaches this problem as a constrained optimization problem and finds adversarial examples with the Lagrangian method, creating a more accurate and powerful adversarial examples \cite{carlini2017towards}. 
Basically, these algorithms require the gradient of each objective function with respect to the inputs. 
Besides gradient-based attacks, gradient-free attacks such as SPSA \cite{uesato2018adversarial}, Square\cite{andriushchenko2020square} and gradient-approximation attacks called Backward Pass Differentiable Approximation (BPDA) \cite{athalye2018obfuscated}, are proposed.
Despite the fact that these methods do not rely on the local gradient, many defenses are circumvented. These methods could be more potent than gradient-based attacks in terms of the applicability.

\subsection{Defense Mechanisms}

 \textbf{Adversarial training}
Adversarial training \cite{goodfellow2014explaining} is training a model with specific adversarial examples. 
The model assumes the potential adversary and minimizes its loss concerning adversarial examples.
Among those adversaries, PGD adversary has been widely chosen, and PGD-trained model shows outstanding robustness against various attack strategies advocated by \cite{madry2017towards}.
Inspired by the success of PGD-trained model, TRADE \cite{zhang2019theoretically} builds a more robust model by modifying the underlying loss function.
Even though adversarial training achieves the decent robustness against various attack strategies, it sometimes requires tremendous training time to converge and may hurt generalization ability harshly. 
It may hinder its applicability in real-world applications.

 \textbf{Pre-processing}
Pre-processing or input transformation is transforming input images for the denoising purpose. 
Some papers \cite{liao2018defense, xie2019feature} focus on the high-level representation to minimize the distance between adversarial and clean examples. \cite{song2017pixeldefend} borrows the idea of a statistical generative model, and the model reconstructs images to be clean-like images using pixelCNN architecture. 
\cite{guo2017countering} applies basic image transformations to remove the underlying adversarial perturbations.
\cite{shen2017ape} and \cite{samangouei2018defense} use GAN architecture to purify the adversarial perturbations by projecting inputs into the legitimate distribution. 
All of these methods insist that their methods are robust against various adversarial examples.
However, those pre-possessing methods are still vulnerable when the adversary can explore the defended model, called the white-box scenario.
Even if the adversary does not know the defense methods, \cite{athalye2018obfuscated} shows that it is still vulnerable to the gradient-approximation attack.
After the gradient-free attacks were introduced by \cite{athalye2018obfuscated} and \cite{uesato2018adversarial}, pre-processing has known to be relatively vulnerable against adversarial examples than adversarial training.
Nevertheless, it is robust against various transfer-based attacks and achieves decent generalization ability.

\section{Method}

We describe the methodology in detail.
This section is organized as follows.
We first present notations used in our work and describe the architecture of the purification model. 
We then explain each loss function and how it contributes to the training.

\subsection{Notation}
Since our training method stems from CycleGAN\cite{zhu2017unpaired}, we have to construct two different domains: \textit{clean data domain} and corresponding \textit{adversarial data domain}.
To be specific, we denote original training samples as \textit{clean} since it was not affected by any attacks.
Let $\mathcal{C}$ and $\mathcal{A}$ denote the domain with clean and adversarial data, respectively.
A target classifier $f$, which is trained with clean data $x$ from $\mathcal{C}$, outputs a classification score over $K$, where $K$ is the number of classes in $\mathcal{C}$.
Adversarial domain $\mathcal{A}$ consists of adversarial data $x_{adv}$ with small perturbation yet effective enough to deceive the target classifier $f$.
An adversary generates adversarial perturbations $\delta$ against the inputs $x \in   [0,1]^{H\times W\times C}$ and combines the perturbations with the inputs to construct the adversarial inputs $x_{adv} = x + \delta$ s.t., $f(x) \neq f(x_{adv})$.
An auxiliary classifier $\eta$ computes the probability to decide where the given inputs come from.
For the purification, a generator $\mathcal{T}$ and a discriminator $\mathcal{D}$ learn the image-to-image translation between $\mathcal{C}$ and $\mathcal{A}$ so that achieving $x' \approx x$ as well as $f(x') \approx f(x)$, where  $x' = \mathcal{T}(x_{adv})$. 
Note that we have two kinds of generator $\mathcal{T}$: $\mathcal{T}_{\mathcal{A} \rightarrow \mathcal{C}}$ and $\mathcal{T}_{\mathcal{C} \rightarrow \mathcal{A}}$.
$\mathcal{T}_{\mathcal{A} \rightarrow \mathcal{C}}$ learns to map the inputs from $\mathcal{A}$ to $\mathcal{C}$, and $\mathcal{T}_{\mathcal{C} \rightarrow \mathcal{A}}$ is vice versa.

\begin{figure}[ht!]
\begin{center}
\includegraphics[width=0.5\textwidth]{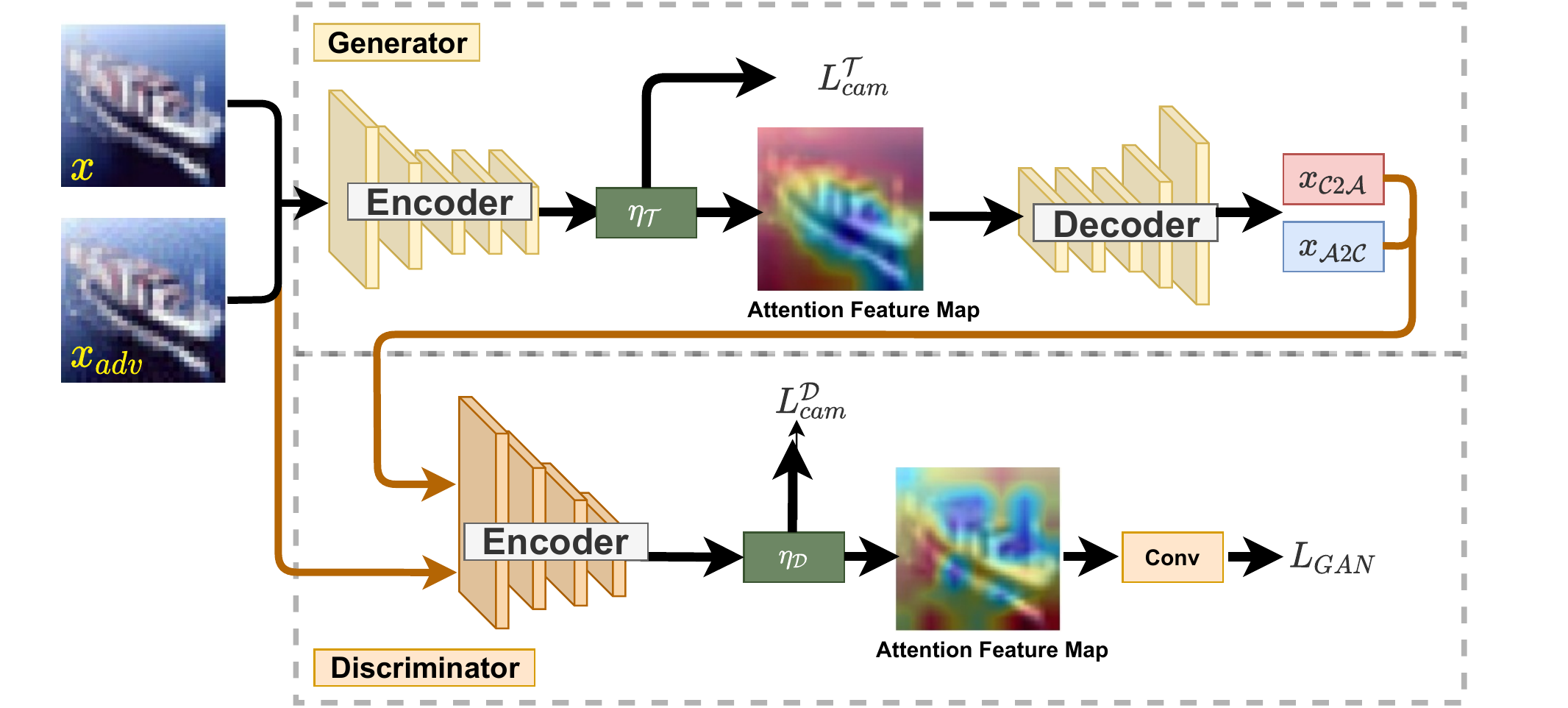}
\end{center}
   \caption{Overview of architecture for generator $\mathcal{T}$ and discriminator $\mathcal{D}$.}
 \label{Arch_G_and_D}
\end{figure}
\subsection{Generative Attentional Purification with Cycle-consistent Learning}
The main idea of our approach is initially originated from the image-to-image translation task\cite{zhu2017unpaired,isola2017image}, where the main goal is training a model to map samples across two different domains correctly.
We apply this concept to mitigate the risk of adversarial attacks. 
As we aim to train $\mathcal{T}_{\mathcal{A} \rightarrow \mathcal{C}}$, we describe the architecture from the perspective of $\mathcal{T}_{\mathcal{A} \rightarrow \mathcal{C}}$.
$\mathcal{T}_{\mathcal{C} \rightarrow \mathcal{A}}$ works in the same manner and is used for cycle-consistent learning to enhance $\mathcal{T}_{\mathcal{A} \rightarrow \mathcal{C}}$. 
The architecture is presented in Figure \ref{Arch_G_and_D}.

\textbf{Generator}
We use the architecture introduced in \cite{kim2019u,zhu2017unpaired} and modify it for our task. 
Basically, the generator $\mathcal{T}_{\mathcal{A} \rightarrow \mathcal{C}}$ consists of mainly three parts: an encoder $\mathrm{E}_{\mathcal{T}_{\mathcal{A}}}$, an auxiliary classifier $\eta_{\mathcal{T}_\mathcal{A}}$, and a decoder $\mathcal{G}_{\mathcal{C}}$. 
The encoder $\mathrm{E}_{\mathcal{T}_\mathcal{A}}$ generates encoded feature maps from a batch of adversarial images $x^B_{adv}$, where $B$ is the size of batch.
The global average pooling layer $\psi$ is then applied to generate spatial averages of feature maps from $\mathrm{E}_{\mathcal{T}_\mathcal{A}}$, to use them as the inputs of the fully connected layer.
Through the auxiliary classifier $\eta_{\mathcal{T}_\mathcal{A}}$, we can compute the probability to determine whether given images (e.g., $x_{adv}$) belong to a specific domain (e.g., $\mathcal{A}$).
As what follows, we leverage weights $w_{\mathcal{T}_\mathcal{A}}$, which represent the importance with respect to $x^B_{adv}$ computed by $\eta_{\mathcal{T}_\mathcal{A}}$, to make the attention map $M^B_{\mathcal{T}_\mathcal{A}}$ for the decoder $\mathcal{G}_{\mathcal{C}}$, i.e., $M^B_{\mathcal{T}_\mathcal{A}} = \sum^n_{k=1}  w^k_{\mathcal{T}_\mathcal{A}} \cdot \psi(E^k_{\mathcal{T}_\mathcal{A}}(x^B_{adv}))$, where $n$ is the number of encoded feature maps.
Considering $\eta_{\mathcal{T}_\mathcal{A}}$ learns domain-specific features, the activated map $M^B_{\mathcal{T}_\mathcal{A}}$ can guide the decoder $\mathcal{G}_{\mathcal{C}}$ to focus on more meaningful regions, leading to encouraging valid source-to-target mapping.
We will describe how $\eta$ works internally in Section \ref{LossFunc_Section}.
The decoder $\mathcal{G}_{\mathcal{C}}$ considers $M^B_{\mathcal{T}_\mathcal{A}}$ as the inputs and reconstructs the purified inputs $x^B_{\mathcal{A}2\mathcal{C}}$  for the target domain $\mathcal{C}$.
Overall, the generator $\mathcal{T}$ takes $x^B_{adv}$ and $x^B$ to purify or perturb given images in the following manner: 
\begin{equation}
\begin{split}
\nonumber
& \scriptstyle  \mathcal{T}_{\mathcal{A} \rightarrow \mathcal{C}}(x^B_{adv}) =  \mathcal{G}_{\mathcal{C}} (\sum_{k=1}^n w_{\mathcal{T}_\mathcal{A}}^{k} \cdot \psi(E^k_{\mathcal{T}_\mathcal{A}}(x^B_{adv}))) = \mathcal{G}_{\mathcal{C}}({M^B_{\mathcal{T}_\mathcal{A}}})=  x^B_{\mathcal{A}2\mathcal{C}} \\
& \scriptstyle\mathcal{T}_{\mathcal{C} \rightarrow \mathcal{A}}(x^B) =  \mathcal{G}_{\mathcal{A}}(\sum_{k=1}^n w_{\mathcal{T}_\mathcal{C}}^{k} \cdot \psi(E^k_{\mathcal{T}_\mathcal{C}}(x^B))) =  \mathcal{G}_{\mathcal{A}}({M^B_{\mathcal{T}_\mathcal{C}}}) = x^B_{\mathcal{C}2\mathcal{A}} \\
& \scriptstyle \mathcal{T}_{\mathcal{A} \rightarrow \mathcal{C}}(x^B_{\mathcal{C}2\mathcal{A}} ) =  \mathcal{G}_{\mathcal{C}}  (\sum^n_{k=1} w_{\mathcal{T}_\mathcal{A}}^{k} \cdot \psi(E^k_{\mathcal{T}_\mathcal{A}} (x^B_{\mathcal{C}2\mathcal{A}}))) = x^B_{\mathcal{C}2\mathcal{A}2\mathcal{C}} \\
& \scriptstyle \mathcal{T}_{\mathcal{C} \rightarrow \mathcal{A}}(x^B_{\mathcal{A}2\mathcal{C}}) =  \mathcal{G}_{\mathcal{A}}(\sum^n_{k=1} w_{\mathcal{T}_\mathcal{C}}^{k} \cdot \psi(E^k_{\mathcal{T}_\mathcal{C}}(x^B_{\mathcal{A}2\mathcal{C}}))) = x^B_{\mathcal{A}2\mathcal{C}2\mathcal{A}}
\end{split}
\end{equation}
Apart from $x^B_{adv}$ and $x^B$, the generator $\mathcal{T}$ takes translated images as inputs since we train the model under cycle-consistent learning.
\begin{figure}[t!]
\begin{center}
\includegraphics[width=0.4\textwidth]{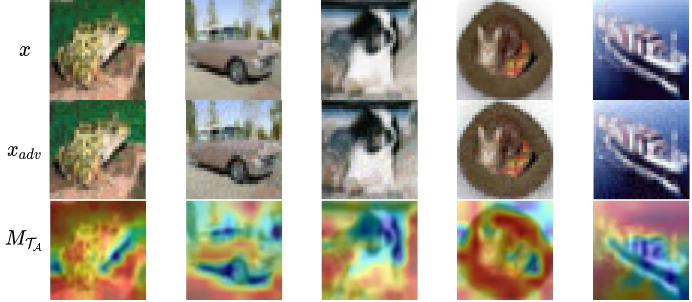}
\end{center}
   \caption{We visualize a few samples of each domain to describe the role of $L_{CAM}$. 
For examples, $x_{adv}$ are fed into $\eta_{\mathcal{T}_{\mathcal{A}}}$ to compute the probability. Once $\eta_{\mathcal{T}_{\mathcal{A}}}$ has learned which regions represent $\mathcal{A}$, its weight will highlight those regions (\textit{i.e., underlying noise}) in the attention map and forward it to the decoder. As a result, the attention map would lead to the decoder to behave in a meaningful way (\textit{i.e., purifying the underlying noise}). }
 \label{Viz_At_map}
\end{figure}

\begin{figure*}[ht!]
\begin{center}
\includegraphics[width=0.8\textwidth]{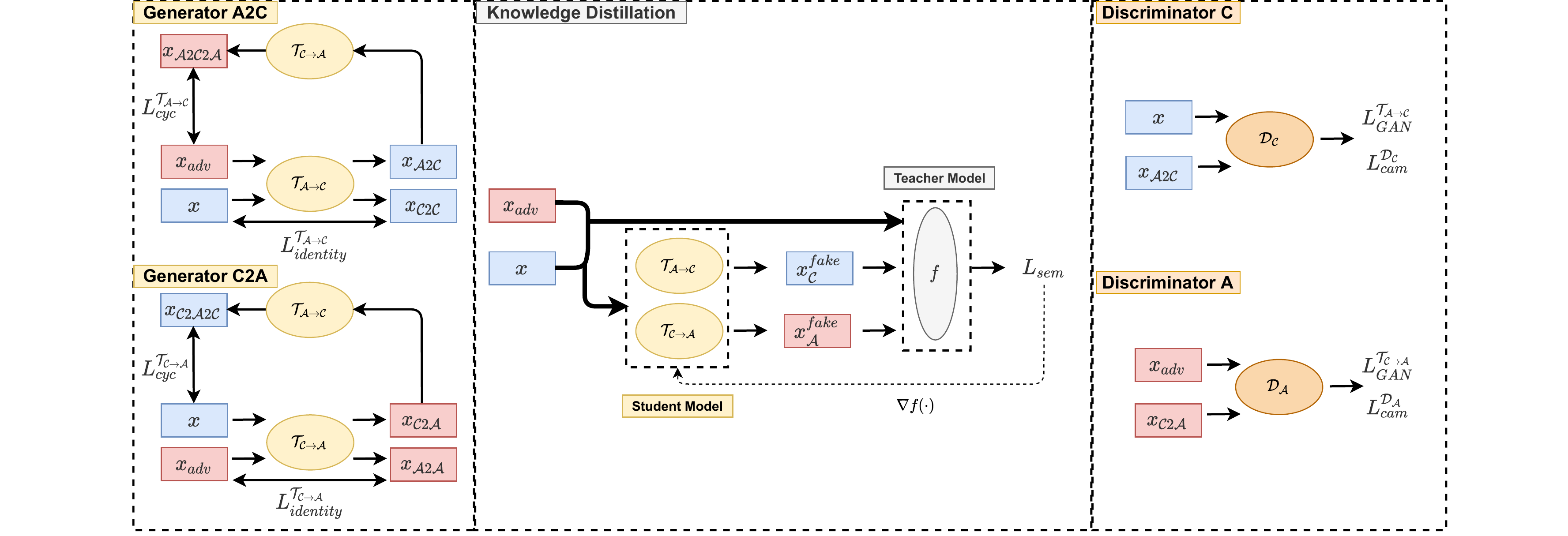}
\end{center}
   \caption{Entire training procedure under cycle-consistent learning. Note that $\mathcal{C}$ is represented using blue, and $\mathcal{A}$ is represented using red color.}
 \label{Training_procedure}
\end{figure*}

\textbf{Discriminator}
$\mathcal{D}_{\mathcal{C}}$ consists of three parts: an encoder $E_{\mathcal{D}_{\mathcal{C}}}$, an auxiliary classifier $\eta_{\mathcal{D}_{\mathcal{C}}}$ and a convolutional classifier. 
The discriminator $\mathcal{D}_{\mathcal{C}}$ is responsible for deciding whether given images are valid or not using the auxiliary classifier $\eta_{\mathcal{D}_{\mathcal{C}}}$ and the convolutional classifier. 
Both classifiers update their loss toward distinguishing whether the given images are original or generated. 
While $\eta_{\mathcal{D}_{\mathcal{C}}}$ takes inputs as encoded feature maps generated from $E_{\mathcal{D}_{\mathcal{C}}}$, the convolutional classifier takes inputs as attention maps $M_{\mathcal{D}_{\mathcal{C}}}$ activated by the weights of $\eta_{\mathcal{D}_{\mathcal{C}}}$.
$\eta_{\mathcal{D}_{\mathcal{C}}}$ will classify $x$ as the images belonging to $\mathcal{C}$ based on the probability, whereas $x_{\mathcal{A}2\mathcal{C}}$ is considered as the images \textit{not} belonging to $\mathcal{C}$. 
$\eta_{\mathcal{D}_{\mathcal{C}}}$ has learned the domain-specific features, so activated regions on $M_{\mathcal{D}_{\mathcal{C}}}$ can be viewed as guided regions for improving the generator $\mathcal{T}$.
We elaborate on how to achieve the mapping between two different domains in the following section.

\subsection{Loss Functions}
\label{LossFunc_Section}
We introduce the objective function of CAP-GAN in detail. 
As we described before, our loss functions are designed with leveraging both pixel-level and  feature-level approach.
We use CycleGAN training procedure for pixel-level consistency ($L_{GAN}, L_{identity}$, and  $L_{cyc}$). 
For feature-level consistency, we use $L_{cam}$, which encourages the model to learn domain-specific features, and $L_{sem}$, which guides the generator to learn the semantic information by distilling the knowledge from the pre-trained model $f$.

\subsubsection{Pixel-level Adaptation}
\paragraph{Adversarial Loss}
To get visually appealing results, we leverage GAN\cite{goodfellow2014generative} training scheme. 
We apply Least Squared GAN (LSGAN) \cite{mao2017least} objective function to ensure that purified images should reside in the vicinity of the original data distribution.
Plus, LSGAN shows more stable learning.
$L_{GAN}$ can be formulated as below:
\begin{equation} 
\begin{split}
\nonumber
L_{GAN} & = L^{\mathcal{T}_{\mathcal{A} \rightarrow \mathcal{C}}}_{GAN} + L^{\mathcal{T}_{\mathcal{C} \rightarrow \mathcal{A}}}_{GAN}  \\
& = \scriptstyle \mathop{\mathbb{E}_{x \sim {\mathcal{C}}}} [(\mathcal{D}_{\small_{\mathcal{C}}} (x))^2 ] + \mathop{\mathbb{E}_{x_{adv} \sim \mathcal{A}}} [(1 - (\mathcal{D}_{\small_{\mathcal{C}}}(\mathcal{T}_{\small_{{\mathcal{A} \rightarrow \mathcal{C}}}}(x_{adv}))))^2 ] \\
& + \scriptstyle \mathop{\mathbb{E}_{x_{adv} \sim {\mathcal{A}}}} [(\mathcal{D}_{\small_{\mathcal{A}}} (x_{adv}))^2 ] +
\mathop{\mathbb{E}_{x \sim \mathcal{C}}} [(1 - (\mathcal{D}_{\small_{\mathcal{A}}}(\mathcal{T}_{\small_{{\mathcal{C} \rightarrow \mathcal{A}}}}(x))))^2 ] 
\end{split}
\end{equation}

\paragraph{Identity Loss \& Cycle Loss} 
Identity loss $L_{identity}$ is able to help get visual satisfaction and plays a role in preserving color composition \cite{zhu2017unpaired}. 
Given images $x$, $\mathcal{T}_{\mathcal{A} \rightarrow \mathcal{C}}(x)$  should not change $x$ because the decoder $\mathcal{G}_\mathcal{C}$ is supposed to learn to decode the images belonging to domain $\mathcal{C}$. 
If the model does not consider $L_{identity}$, $\mathcal{T}_{\mathcal{A} \rightarrow \mathcal{C}}$ will change non-essential points and affect both adversarial and cycle-consistency loss in unexpected ways \cite{zhu2017unpaired}.
\begin{equation} 
\begin{split}
\nonumber
L_{identity} & = L^{\mathcal{T}_{\mathcal{A} \rightarrow \mathcal{C}}}_{identity} + L^{\mathcal{T}_{\mathcal{C} \rightarrow \mathcal{A}}}_{identity}  \\
& = \scriptstyle \mathop{\mathbb{E}_{x \sim \mathcal{C}}||\mathcal{T}_{\mathcal{A} \rightarrow \mathcal{C}}(x) - x} ||_1 \\
& + \scriptstyle \mathop{\mathbb{E}_{x_{adv} \sim \mathcal{A}}||\mathcal{T}_{\mathcal{C} \rightarrow \mathcal{A}}(x_{adv}) - x_{adv}} ||_1
\end{split}
\label{Puri:ID}
\end{equation}
Cycle-consistency loss $L_{cyc}$ is responsible for preserving the semantic information.
Based on the objective function $L_{cyc}$, transformed images should be able to revert back to source images. 
Such a constraint would tighten the relationship between two different domains and ensure that transformed images include more semantic information like structural features or local textures.
\begin{equation} 
\begin{split}
\nonumber
L_{cyc} & = L^{\mathcal{T}_{\mathcal{A} \rightarrow \mathcal{C}}}_{cyc} + L^{\mathcal{T}_{\mathcal{C} \rightarrow \mathcal{A}}}_{cyc}  \\
& = \scriptstyle \mathop{\mathbb{E}_{x_{adv} \sim \mathcal{A}}} || \mathcal{T}_{\mathcal{}{C} \rightarrow \mathcal{A}} (\mathcal{T_{\mathcal{A} \rightarrow \mathcal{C}}}({x_{adv}})) - x_{adv}||_1 \\
& +\scriptstyle \mathop{\mathbb{E}_{x \sim \mathcal{C}}} || \mathcal{T}_{\mathcal{A} \rightarrow \mathcal{C}} (\mathcal{T_{\mathcal{C} \rightarrow \mathcal{A}}}(x)) - x||_1
\label{Puri:Cycle}
\end{split}
\end{equation}
$L_{identity}$ and  $L_{cyc}$ contribute to mapping the two different distributions from the source domain to the target domain.
We use $L_1$ norm to compute each loss function because it shows better results than $L_2$ norm in terms of both the accuracy and the quality of images.

\subsubsection{Feature-level Adaptation}
\paragraph{Cam Loss}
Motivated by \cite{zhou2016learning} and \cite{kim2019u}, we apply an attention module in the intermediate layer of the generator.
Class activation map (CAM) is originally designed to explain how the model works internally. 
We borrow this idea to guide the generator to figure out which regions need to be improved.
Instead of the classification, we make the auxiliary classifier learning domain-specific features.
By doing so, $\eta_{\mathcal{D}}$ can figure out whether given inputs include domain-specific features or not, and its weights are able to represent the domain properties. 
We then generate the attention map over the inputs using weights of $\eta_{\mathcal{D}}$. 
According to the attention map generated by $\eta_{\mathcal{D}}$, $\mathcal{T}$ tries to figure out which regions are important.
For computing $L^{\mathcal{D}}_{cam}$, we use LSGAN training loss as $L_{GAN}$ used.
\begin{equation} 
\begin{split}
\nonumber
L^{\mathcal{D}}_{cam} & = L^{\mathcal{D}_{\mathcal{A}}}_{cam} + L^{\mathcal{D}_{\mathcal{C}}}_{cam}  \\
& = \scriptstyle\mathop{\mathbb{E}_{x_{adv} \sim \mathcal{A}}}[(\eta_{\mathcal{D}_\mathcal{A}}(x_{adv}))^2]  + {\mathop{\mathbb{E}_{x \sim \mathcal{C}}}[(1-\eta_{\mathcal{D}_\mathcal{A}}(\mathcal{T}_{\mathcal{C} \rightarrow \mathcal{A}}(x)))^2}] \\
& \scriptstyle \,  + {\mathop{\mathbb{E}_{x \sim \mathcal{C}}[(\eta_{\mathcal{D}_\mathcal{C}}(x))^2}]}  + \mathop{\mathbb{E}_{x_{adv} \sim \mathcal{A}}}[(1-\eta_{\mathcal{D}_\mathcal{C}}(\mathcal{T}_{\mathcal{A} \rightarrow \mathcal{C}}(x_{adv})))^2]    \\
L^{\mathcal{T}}_{cam} & = L^{\mathcal{T}_{\mathcal{A} \rightarrow \mathcal{C}}}_{cam} + L^{\mathcal{T}_{\mathcal{C} \rightarrow \mathcal{A}}}_{cam}  \\
& = \scriptstyle -\mathop{\mathbb{E}_{x_{adv} \sim \mathcal{A}}}[log(\eta_{\mathcal{T}_\mathcal{A}}(x_{adv}))] + 
\mathop{\mathbb{E}_{x \sim \mathcal{C}}}[log(1-\eta_{\mathcal{T}_\mathcal{A}}(x))] \\
& - \scriptstyle\mathop{\mathbb{E}_{x \sim \mathcal{C}}}[log(\eta_{\mathcal{T}_\mathcal{C}}(x))] + 
\mathop{\mathbb{E}_{x_{adv} \sim \mathcal{A}}}[log(1-\eta_{\mathcal{T}_\mathcal{C}}(\mathcal{T}_{\mathcal{C \rightarrow \mathcal{A}}}(x_{adv})))] \\
L_{cam} & = L^{\mathcal{D}}_{cam}  + L^{\mathcal{T}}_{cam}
\end{split}
\label{Puri:CAM_D}
\end{equation}
Besides GAN training in $L_{cam}$, we additionally put the term to encourage the generator $\mathcal{T}$ to recognize the source domain since it turns out that the discriminator easily overwhelms the generator in training.
We perform binary classification for the auxiliary classifier $\eta_{\mathcal{T}}$ to decide where the images come from.
For instance, the probability computed from $\eta_{\mathcal{T}}$, which is the auxiliary classifier learning the domain $\mathcal{A}$, should be closer to 1 if the inputs come from $\mathcal{A}$, as opposed to 0 if the inputs come from $\mathcal{C}$.
We provide visual intuition about $L_{cam}$ in Figure \ref{Viz_At_map}.

\paragraph{Semantic Loss}
CycleGAN is not designed for the classification purpose, but for visually appealing images in the pixel-level consistency, such as the image-to-image translation \cite{zhu2017unpaired,kim2019u}.
Even if we can get visually similar results at the pixel-level, it does not mean that low-level appearance consistency can bring the semantic consistency like the classification accuracy (See Figure \ref{tsne_viz}) \cite{hoffman2018cycada}.
With these perspectives, we design the generator to consider the semantic consistency by distilling the knowledge from the pre-trained model $f$.
In other words, we impose the semantic-level similarity between generated and original data in both domains.

We combine the knowledge of $f$ into $\mathcal{T}$ for providing better understanding of the feature-level consistency.
That is, we conjecture that well-trained model $f$ can convey useful information to $\mathcal{T}$, including the weakness of $f$ against the domain $\mathcal{A}$.
Instead of using ground-truth labels, we use soft-targets \cite{hinton2015distilling}, where the class-wise scores are estimated based on a softmax function, by projecting the images on the pre-trained embedding space of $f$.
We adopt the concept of cycle-consistent learning to the semantic loss landscape as well.
We compute the logit values, where $z^{fake}_{\mathcal{C}} = f(x^{fake}_{\mathcal{C}})$ and $z^{fake}_{\mathcal{A}} = f(x^{fake}_{\mathcal{A}})$, under cycle-consistent learning.
$x^{fake}_{\mathcal{C}}$and $x^{fake}_{\mathcal{A}}$ represent the generated images with respect to the domain $\mathcal{C}$ and $\mathcal{A}$ respectively, i.e., ${\{{x_{\mathcal{A}2\mathcal{C}}, x_{\mathcal{C}2\mathcal{C}}, x_{\mathcal{C}2\mathcal{A}2\mathcal{C}}\}}}$ and $\{ {x_{\mathcal{C}2\mathcal{A}}, x_{\mathcal{A}2\mathcal{A}}, x_{\mathcal{A}2\mathcal{C}2\mathcal{A}}\}}$.
We match them to the soft-targets $z^{real}_{\mathcal{C}} = f(x)$ and $z^{real}_{\mathcal{A}} = f(x_{adv})$, respectively. 
We can generally formulate the cycle-consistent semantic term as below:
\begin{equation} 
\begin{split}
\nonumber
L_{sem} & = T^2 \cdot \mathrm{KL}(p(z^{real}_{\mathcal{C}}, T), p(z^{fake}_{\mathcal{C}},T)) \\
& + T^2 \cdot \mathrm{KL}(p(z^{real}_{\mathcal{A}}, T), p(z^{fake}_{\mathcal{A}},T))
\end{split}
\end{equation}
, where $\mathrm{KL}$ is Kullback-Leibler divergence, and $p$ is the class-wise scores computed by a softmax function with a scaling temperature $T$, $p(z_{i}, T) =  \frac{exp(z_{i}/T) }{\sum_{j} exp(z_{j}/T) }$.
$T$ is in charge of smoothing the distribution over the classes when logit $z$ is given. 
We will discuss how $T$ affects the performance in Section \ref{Loss_Functions_AS}.

\subsubsection{Entire Objective Function}
To utilize both pixel-level and feature-level adaptation, CAP-GAN objective function is a weighted average of the loss functions.
$\alpha$ is a balancing parameter to adjust the weights of $L_{pixel}$ and $L_{feature}$ contribution, where $L_{pixel}$ is $L_{GAN} + L_{identity}+ L_{cyc}$ and $L_{feature}$ is $L_{cam} + L_{sem}$.
The pixel-level loss is conducive to preserve the structural shapes or the texture features, whereas the feature-level loss encourages the model to decode the images in a semantically meaningful way.
We depict the entire training procedure in Figure \ref{Training_procedure}.
\begin{equation}
\nonumber
L_{CAP}  =  \alpha \cdot  L_{pixel}  + (1-\alpha) \cdot L_{feature}
\end{equation}

\section{Experiments}

We conduct the experiments to compare the robustness with other state-of-the-art defense methods.
Note that we measure the robustness by the classification accuracy under $l_\infty$-ball attacks, except for $l_{2}$ CW attack with 1000 steps.
We evaluate the robustness under both black-box and white-box settings with $\epsilon=8$.
We then provide an ablation study to show how the performance can be affected by each hyper-parameter in Section \ref{Parameters_AS}.

\subsection{Experiment Details}
We train the proposed model on CIFAR-10 dataset with a single RTX TITAN 2080. 
We use a simple ResNet model as a victim model following \cite{madry2017towards}.
We train the model up to 200 epochs using Adam optimizer ($\beta$ = (0.5,0.999)) and set batch size as 128. 
We use the initial learning rate 1e-4 and adopt cosine annealing learning rate decay for a stable training.
To construct the adversarial domain $\mathcal{A}$, we use FGSM ($\epsilon=8$) attack \cite{goodfellow2014explaining}.
We set the $\alpha=0.7$ and $T=10.0$ to train the model.
For a fair comparison, we implement all the defenses using the same codebase.

\subsection{Evaluation on Black-Box Attacks}
A black-box attack is an attack where an adversary has limited knowledge of the defense. 
This section assumes that a black-box adversary is not aware of the defense but has the knowledge of the target model $f$.
We evaluate the robustness of our defense against various attack strategies.
We follow the black-box assumption introduced in \cite{liu2016delving} and use the same attack settings with \cite{madry2017towards}.
Specifically, we provide the architecture of the target model $f$ to the adversary and train it with the same training data.
With fully trained surrogate model $f'$, the adversary then generates transferable adversarial examples using gradient-based attacks such as FGSM, PGD$_{n}$ and CW($\kappa=20$) strategy, where $n$ is the number of step, and $\kappa$ is a confidence.
Plus, to endanger the defense under a more harsh transfer-attack, we apply the momentum iterative FGSM, called MI-FGSM  \cite{dong2018boosting}. We use 20 iterative steps for MI-FGSM.

As shown in Table \ref{BB_CIFAR}, CAP-GAN shows promising results against transfer-based attacks and outperforms other pre-processing based methods, including the adversarial training(AT) model. 
We trained AT model using PGD$_{7}$ adversary following \cite{madry2017towards}.
Interestingly, FGSM works better than its iterative attack PGD$_{n}$ in APE, HGD, and CAP-GAN. 
Such an observation is on par with the tendency, where the gradient masking defense makes the iterative attacks getting stuck in a local minimum \cite{athalye2018obfuscated}. 
THM \cite{buckman2018thermometer} achieves superb results in the first-step attack, such as FGSM attack, whereas it is relatively vulnerable to the iterative attacks, such as MI-FGSM and PGD. 
Especially, HGD obtains similar performance with CAP-GAN.

\begin{figure}[ht]
\begin{tabular}{lccc}
	\includegraphics[width =4.3cm]{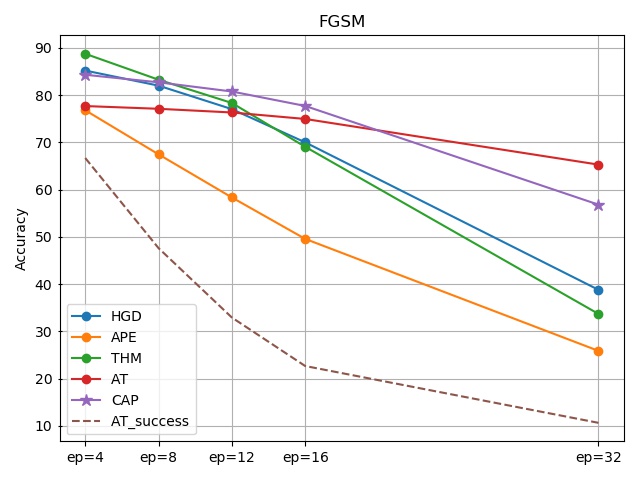} 
	\includegraphics[width =4.3cm]{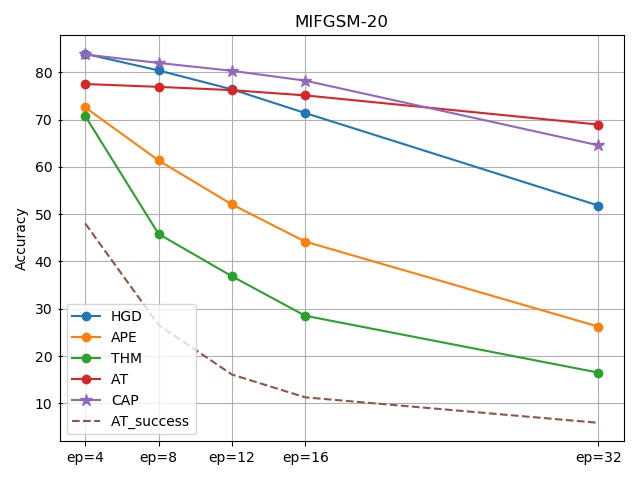} \\
	\includegraphics[width =4.3cm]{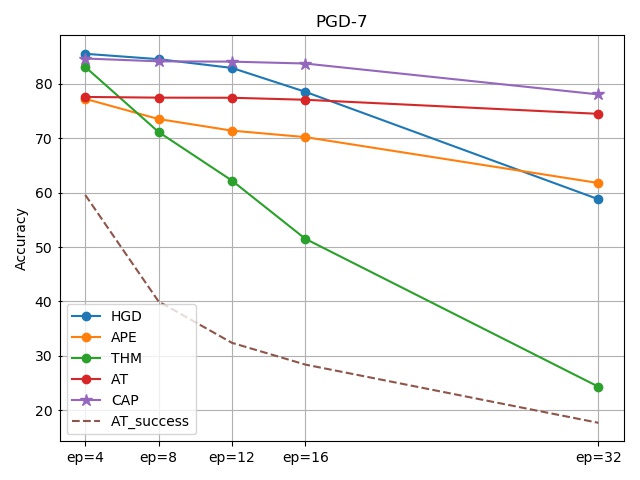} 
	\includegraphics[width =4.3cm]{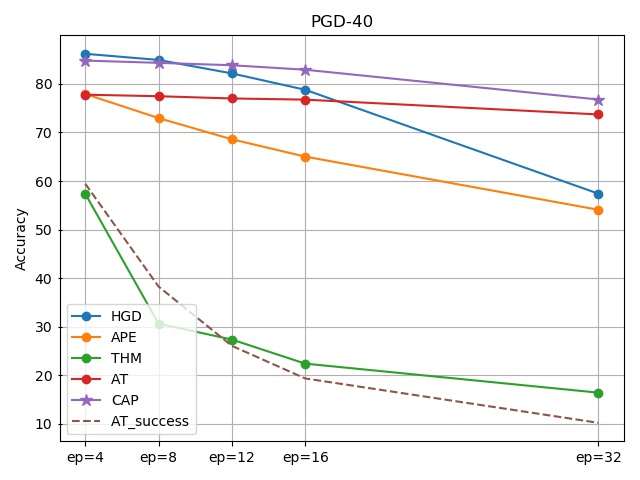} 
\end{tabular}
\caption{\textbf{Black-box results for CIFAR-10:} We vary the magnitude of the perturbation to examine the adversarial robustness. In all defenses, the degradation occurs as the magnitude of perturbation increases. We also provide the attack success rate (AT\_success) to examine the capacity of the adversary. }

\label{fig:bb_graph}
\end{figure}

Even though we use the surrogate model $f'$ with the same training settings, it cannot ensure that $x_{adv}$ generated from $f'$ has transferability to $f$ \cite{uesato2018adversarial}.
We thus measured the attack success rate, which is referred to as the rate of misclassification against the target model $f$ when the attack strategy is applied, to judge whether the attack has enough capacity to deceive the model.
We observed that generating the adversarial examples with $\epsilon=8$ is insufficient to assume the strong adversary. 
We conjectured that the surrogate model $f'$ might not have a similar gradient direction of $f$, implying that crafting the adversarial examples with the gradient direction of $f'$ requires a large magnitude of perturbation to achieve a high attack success rate \cite{liu2016delving}.
Therefore, we decided to evaluate the defenses under various magnitudes of the perturbation $\epsilon=[4,8,12,16,32]$ to argue more clear robustness against various attack strategies.
The results are presented in Figure \ref{fig:bb_graph}. 
\begin{table}[ht!]
\centering
\renewcommand{\arraystretch}{0.8}
\begin{tabular}{@{}lccccc@{}}
\toprule
 \multicolumn{1}{c}{\multirow{2}{*}{\textbf{Method}}} &  \multicolumn{5}{c}{\textbf{Accuracy(\%)}}                \\ \cmidrule{2-6}
   &  {\textbf{Clean}} &  {\textbf{FGSM}} &  {\textbf{MI-FGSM$_{20}$}} & {\textbf{PGD$_{7/40}$}} &   {\textbf{CW}} 
\\  \midrule
{JPEG \cite{dziugaite2016study}}  & 74.76 & 59.53 & 54.86 & 67.1/66.03 &  71.06 \\
{THM\cite{buckman2018thermometer}}    & \textbf{89.96} & \textbf{83.25} & 45.81 & 71.17/30.63& -    \\
{APE\cite{shen2017ape}}  & 84.15 & 67.52& 61.54 & 73.21/72.32    & 80.3 \\
{HGD \cite{liao2018defense}}   & 87.68  & 82.35 & 80.37  & \textbf{84.55}/84.39     & \textbf{85.36} \\
{AT\cite{madry2017towards}}            & 78.1  & 77.1  & 76.9  & 77.49/77.46   & 78.36\\ \midrule
{\textbf{CAP-GAN}} & 85.62 & 82.73 & \textbf{81.96} & 84.22/\textbf{84.43} & 84.78 \\  \bottomrule
\end{tabular}
\caption{\textbf{Black-Box results for CIFAR-10}: We train the surrogate model $f'$ to generate transferable adversarial examples. Before feeding the adversarial examples into the target classifier $f$, those examples are projected into $\mathcal{T}_{\mathcal{A} \rightarrow \mathcal{C}}$.}
\label{BB_CIFAR}
\end{table}

\subsection{Extensive Black-Box Attacks}

Providing $\mathcal{T}_{\mathcal{A} \rightarrow \mathcal{C}}$ for the adversary can cast the light for understanding  effectiveness of the proposed defense.
Besides gradient-based attacks, we also apply the score-based attack, SPSA \cite{uesato2018adversarial} and Square attack \cite{andriushchenko2020square} to launch the strong black-box attack.
In SPSA and Square attack, the adversary is allowed to directly access the outputs of the defended model $f(\mathcal{T}_{\mathcal{A} \rightarrow \mathcal{C}}(\cdot))$.
Since those gradient-free attacks do not rely on the local gradient information, we can provide a different aspect of the robustness.
We set batch size as 2048 for SPSA to make it more potent.
We mount Square attack varying the restart by 1 and 5. 

Apart from gradient-free attacks, we also consider an adaptive adversary who exploits the fundamental weakness of the proposed model.
We assume the adaptive adversary as BPDA-I attack, where the adversary crafts the corresponding adversarial examples using PGD$_{n}$  but backpropagating the gradients with the identity mapping.
According to the experimental results from \cite{athalye2018obfuscated}, many pre-processing based methods \cite{song2017pixeldefend,guo2017countering,buckman2018thermometer} are circumvented under BPDA-I attack.
Note that BPDA attack is typically applied when the defense is non-differentiable, but we utilize this attack scheme to circumvent our defense as  our model is designed to fulfill  $\mathcal{T}_{\mathcal{A} \rightarrow \mathcal{C}}(x_{adv}) \approx x$ and $\mathcal{T}_{\mathcal{A} \rightarrow \mathcal{C}}(x) \approx x$ simultaneously.
As advocated by \cite{athalye2018obfuscated}, BPDA attack can be applied to verify the vulnerability of the model even if the model is differentiable.
We thus assume that the strong black-box adversary exploits the internal weakness $\mathcal{T}_{\mathcal{A} \rightarrow \mathcal{C}}(x) \approx x$  by replacing the backward pass with the identity function to approximate the gradients over the inputs i.e., leveraging the concept that $\nabla_{x} \mathcal{T}(x) \approx \nabla_{x}x =1$, where $x \in \{{\mathcal{C}, \mathcal{A}}\}$.
The experimental results are presented in Table \ref{GreyBox_Attack}.
Our proposed model outperforms other methods across all attacks.
\begin{table}[!ht]
\centering
\renewcommand{\arraystretch}{0.6}
\begin{tabular}{@{}llll@{}}
\toprule
\multicolumn{1}{c}{\multirow{2}{*}{\textbf{Method}}} &  \multicolumn{3}{c}{\textbf{Accuracy(\%)}}                \\ \cmidrule{2-4}
  & \textbf{SPSA} & \textbf{Square$_{r=1/5}$} & \textbf{BPDA-I$_{40}$}   \\ \midrule
 JPEG \cite{dziugaite2016study} & 74.9  & 16.03/7.35 & 0.59 \\
APE \cite{shen2017ape}  & 51.33 & 33.57/24.03 &  0.09\\ 
HGD \cite{liao2018defense}  & 31.26&  23.88/14.26 & 71.78 \\ 
\midrule
\textbf{CAP-GAN} 	 & \textbf{79.28} & \textbf{53.07/43.6} & \textbf{78.27}\\\bottomrule
\end{tabular}
\caption{\textbf{Extensive Black-Box results for CIFAR-10}: We provide the defense $\mathcal{T}_{\mathcal{A} \rightarrow \mathcal{C}}$ to examine the robustness against the gradient-free and the adaptive attack. That is, the adversary can access the outputs of $f(\mathcal{T}_{\mathcal{A} \rightarrow \mathcal{C}}(\cdot))$.}
\label{GreyBox_Attack}
\end{table}

\subsection{Evaluation on White-Box Attacks}
A white-box adversary has full knowledge of the target model and defended model to launch adversarial attacks.
We consider two kinds of white-box settings: conventional white-box setting following \cite{liao2018defense,shen2017ape} and \cite{song2017pixeldefend}, where we provide the target model $f$ to the adversary, and end-to-end attack, where we allow the adversary to directly compute the gradient with respect to the combined model $f(\mathcal{T}_{\mathcal{A} \rightarrow \mathcal{C}}(\cdot))$ following \cite{zantedeschi2017efficient} and \cite{athalye2018obfuscated}.

\textbf{White-box Attack}
In Table \ref{WB_CIFAR}, the target model reaches a nearly zero accuracy when we apply the white-box iterative attacks even with $\epsilon=8$. 
The accuracy of most methods significantly dropped compared to the black-box setting in Table \ref{BB_CIFAR}, whereas HGD and CAP-GAN show relatively consistent results.
The intuition behind both models is using the feature-level consistency when optimizing the denoising loss function. 
However, models that try to align only pixel-level distribution cannot achieve the consistent robustness.
This observation is on par with our assumption described in Section \ref{Introduction}, where the pixel-level consistency cannot assure the feature-level consistency.
As observed by \cite{liao2018defense,goodfellow2014explaining}, the residual perturbations continuously incur the error amplification effect in several intermediate layers, resulting in the misclassification at the prediction layer.
The feature-level consistency additionally aligns the shift caused by the residual perturbations at the prediction layer, so it could help to achieve reasonable purification and leads to consistent performance.

 \begin{table}[t]
\centering
\renewcommand{\arraystretch}{0.8}
\begin{tabular}{@{}lccccc@{}}
\toprule
\multicolumn{1}{c}{\multirow{2}{*}{\textbf{Method}}} &  \multicolumn{5}{c}{\textbf{Accuracy(\%)}}                \\ \cmidrule{2-6}
   &  {\textbf{Clean}} &  {\textbf{FGSM}} &  {\textbf{MI-FGSM$_{20}$}} & {\textbf{PGD$_{7/40}$}} &   {\textbf{CW}} 
\\  \midrule
ResNet & 92.8 & 28.81 & 0.94  & 0.56/0.01 & 0.28 \\
{JPEG \cite{dziugaite2016study}}  & 74.76 & 47.37 & 34.33 & 53.07/50.7  &  68.23  \\
{THM\cite{buckman2018thermometer}}    & \textbf{89.96} & 66.25& 20.01 & 53.82/10.74 & -    \\
{APE\cite{shen2017ape}}  & 84.15 & 49.53& 30.76 &54.94/55.32   & 72.07 \\
{HGD \cite{liao2018defense}}   & 87.68  & 81.55 & 78.95 & \textbf{84.65/84.77}    & \textbf{86.07} \\ \midrule
{\textbf{CAP-GAN}} & 85.62 & \textbf{82.48} & \textbf{81.45} & 84.07/83.93 & 84.9\\  \bottomrule
\end{tabular}
\caption{\textbf{White-Box results for CIFAR-10}: We provide trained weights of the target model $f$ to generate strong adversarial examples.}
\label{WB_CIFAR}
\end{table}

\textbf{End-to-End Attack}
Since APE and HGD are differentiable, we pick those methods for the end-to-end attack to compare the robustness.
Table \ref{WhiteBox_Attack} shows the robustness under the end-to-end white-box setting. 
As expected, gradient-based attack is sufficient to circumvent APE and HGD.
In contrast, adversarial training(AT) shows relatively consistent performance across all attack strategies.
Overall, CAP-GAN surpasses the other differentiable pre-processing defenses, but is relatively less robust than AT in the end-to-end attack setting. 

\begin{table}[!ht]
\centering
\renewcommand{\arraystretch}{0.6}
\begin{tabular}{@{}lcccc@{}}
\toprule
\multicolumn{1}{c}{\multirow{2}{*}{\textbf{Method}}} &  \multicolumn{4}{c}{\textbf{Accuracy(\%)}}                \\ \cmidrule{2-5}
   &  {\textbf{Clean}} &  {\textbf{FGSM}} &  {\textbf{MI-FGSM$_{20}$}} & {\textbf{PGD$_{7/40}$}}
\\  \midrule
APE \cite{shen2017ape}  & 84.15 & 18.96 & 1.13 & 0.58/0.02\\ 
HGD \cite{liao2018defense}  & \textbf{87.68}&  26.6 & 0.63 & 0.67/0.01 \\ 
AT \cite{madry2017towards}  & 78.1 &  \textbf{58.46}  & \textbf{49.54} & \textbf{48.09/45.22} \\ \midrule
\textbf{CAP-GAN} 	 & 85.62 & 53.11 &19.24 & 25.17/6.58\\\bottomrule
\end{tabular}
\caption{\textbf{End-to-End Attack results for CIFAR-10}: We provide the defense $\mathcal{T}_{\mathcal{A} \rightarrow \mathcal{C}}$ to examine the robustness against gradient-based attacks. That is, the adversary crafts the adversarial examples using $f(\mathcal{T}_{\mathcal{A} \rightarrow \mathcal{C}}(\cdot))$.}
\label{WhiteBox_Attack}
\end{table}

\subsection{Ablation Study}
\textbf{Parameter Settings}
\label{Parameters_AS}
We perform a simple ablation study to explain how each parameter affects the performance.
CAP-GAN has two parameters: $\alpha$ and $T$.
$\alpha$ adjusts the contribution between pixel-level and feature-level consistency.
In $L_{sem}$, $T$ is a scaling parameter to smoothen the logits before feeding them into the softmax function.
If we use a high value of $T$, the logit space would be smoother, leading to generating soft-target scores over the inputs.
According to Table \ref{alpha_and_T}, we experimentally found that CAP-GAN obtains the best result with $\alpha=0.7$ and $T=10.0$.
One can note that the accuracy under BPDA-I attack dropped as $\alpha$ increases, implying that $\nabla_{x} \mathcal{T}(x) \approx \nabla_{x}x$ is closer to 1.
This observation can provide an evidence for our assumption in which BPDA-I attack could be an adaptive adversary against CAP-GAN.
\begin{table}[hbt!]
\renewcommand{\arraystretch}{0.6}
\begin{tabular}{@{}lcccc@{}}
\toprule
\multicolumn{1}{c}{\multirow{2}{*}{\textbf{Model}}} &  \multicolumn{4}{c}{\textbf{Accuracy(\%)}}                \\ \cmidrule{2-5}
 & \textbf{Clean} & \textbf{MI-FGSM$_{20}$} & \textbf{PGD$_{7/40}$} & \textbf{BPDA-I$_{40}$}  \\ \midrule
 $\alpha, T = 0.7, 1.0$ & 81.53 & 78.47  & 80.25/80.24 &  71.35 \\ 
 $\alpha, T = 0.7, 5.0$ &	84.81	& 80.97 & 83.39/83.29 & \textbf{79.19}\\ 
 $\alpha, T = 0.7, 10.0$	& \textbf{85.62} & \textbf{81.96}& \textbf{84.22/84.43} & 78.27 \\ \midrule \midrule
 $\alpha, T = 0.5, 10.0$ & 83.96 & 80.94  & 82.67/82.67 &  \textbf{82.71} \\ 
 $\alpha, T = 0.7, 10.0$ &	\textbf{85.62}	& \textbf{81.96} & \textbf{84.22/84.43} & 78.27\\ 
 $\alpha, T = 0.9, 10.0$	& 85.3 & 81.43 & 83.76/83.76 & 63.19 \\ \bottomrule
\end{tabular}
\caption{\textbf{Ablation study for balancing parameters:} We performed black-box attacks to evaluate each setting. }
\label{alpha_and_T}
\end{table}

\textbf{Loss Functions}
\label{Loss_Functions_AS}
We introduce how each loss function contributes internally. 
Besides the pixel-level alignment with $L_{pixel}$, we explicitly add the feature-level terms $L_{feature}=L_{cam}+L_{sem}$ to tune the generator.
$L_{cam}$ helps the model to understand domain-specific features, and $L_{sem}$ conveys distilled knowledge from the pre-trained model to guide $\mathcal{T}$.
Furthermore, we adopt CycleGAN training scheme for enhancing the expected result.
As shown in Table \ref{Loss_Term}, the feature-level terms lead to the improvement in the adversarial robustness across all attacks. 
We initially conjecture that $L_{cam}$ enhances the purification by encouraging the valid source to target mapping, but rather degrades the performance. 
When we combine the model with the semantic term $L_{sem}$, however, the model significantly improves its capacity.
We also found that simply combining the two loss terms would not derive the positive effects. 
At this point, cycle-consistent learning is an important factor in deriving the expected improvement.
This is because $\mathcal{T}_{\mathcal{A} \rightarrow \mathcal{C}}$ takes the inputs from not only domain $\mathcal{A}$ but also $\mathcal{T}_{\mathcal{C} \rightarrow \mathcal{A}}(x) = x_{\mathcal{C}2\mathcal{A}}$ to fulfill cycle-consistency. 
Cycle-consistent learning can regularize the entire training with more diverse inputs.
Thus, additional inputs convey the meaningful hidden information to the model, encouraging the model to understand the underlying relationship between two different domains.

\begin{table}[ht]
\renewcommand{\arraystretch}{0.6}
\begin{tabular}{@{}lcccc@{}}
\toprule
\multicolumn{1}{c}{\multirow{2}{*}{\textbf{Model}}} &  \multicolumn{4}{c}{\textbf{Accuracy(\%)}}                \\ \cmidrule{2-5}
& \textbf{Clean} & \textbf{MI-FGSM$_{20}$} & \textbf{PGD$_{7/40}$} & \textbf{BPDA-I$_{40}$}  \\ \midrule
$L_{pixel}$ & \textbf{90.5} & 57.03 & 77.7/77.67 &  0.34 \\ 
$L_{pixel}$ + $L_{cam}$  &	88.0	& 50.9 & 75.13/75.05 & 0.2 \\ 
$L_{pixel}$ + $L_{sem}$ & 83.53 & 78.79 & 81.64/81.75 & 38.71 \\ \midrule
CAP-GAN \scriptsize w/o Cyc & 82.39  & 78.58 & 80.52/80.67 & 58.18 \\
CAP-GAN & 85.62 & \textbf{81.96} & \textbf{84.22/84.43} &\textbf{78.27} \\
 \bottomrule
\end{tabular}
\caption{\textbf{Ablation study for loss terms:} $L_{pixel}$ means the pixel-level loss terms which consist of $L_{GAN}, L_{identity}$, and $L_{cyc}$. CAP-GAN w/o Cyc indicates the model without cycle-consistent learning in both pixel-level and feature-level.}
\label{Loss_Term}
\end{table}

\section{Conclusion}
We propose CAP-GAN to purify the adversarial examples leveraging both pixel-level and feature-level adaptation. 
While the previous approaches construct task-specific functions to have feature-level consistency implicitly, we explicitly introduce the additional guided terms to achieve the feature-level consistency.
Even more, we apply cycle-consistent learning to regularize the model to be more effective.
CAP-GAN achieves the adversarial robustness against various attack strategies and outperforms other pre-processing methods under both black-box and white-box settings.

\section*{Acknowledgement}
This research was supported by the MSIT(Ministry of Science and ICT), Korea, under the Grand Information Technology Research Center support program(IITP-2021-2020-0-01489) supervised by the IITP(Institute for Information \& communications Technology Planning \& Evaluation), and the Technology Innovation Program(or Industrial Strategic Technology development Program, 2000682, Development of Automated Driving Systems and Evaluation)funded by the Ministry of Trade, Industry \& Energy(MOTIE,Korea).

\bibliography{egbib}
\end{document}